\documentclass[conference]{IEEEtran}
\IEEEoverridecommandlockouts
\usepackage{cite}
\usepackage{amsmath,amssymb,amsfonts}
\usepackage{microtype}

\usepackage{algorithm2e}
\usepackage{graphicx}
\usepackage{svg}
\usepackage{pdfpages}  
\usepackage{subcaption}  

\def\BibTeX{{\rm B\kern-.05em{\sc i\kern-.025em b}\kern-.08em
    T\kern-.1667em\lower.7ex\hbox{E}\kern-.125emX}}
\begin{document}

\title{VisionGRU: A Linear-Complexity RNN Model for Efficient Image Analysis}

\author{
Shicheng Yin\textsuperscript{1}\textsuperscript{*}\quad Kaixuan Yin\textsuperscript{1}\textsuperscript{*} \quad 
Weixing Chen\textsuperscript{1} \quad Enbo Huang\textsuperscript{2} \quad Yang Liu \textsuperscript{1}\textsuperscript{$\diamond$}\\
{\normalsize
\textsuperscript{1}Sun Yat-sen University \quad       
\textsuperscript{2}Nanning Normal University \
}\\
\tt\small \{yinshch5, yinkx3, chenwx228\}@mail2.sysu.edu.cn, hebscitech@outlook.com, liuy856@mail.sysu.edu.cn\\
}

\maketitle

\begin{abstract}
Convolutional Neural Networks (CNNs) and Vision Transformers (ViTs) are two dominant models for image analysis. While CNNs excel at extracting multi-scale features and ViTs effectively capture global dependencies, both suffer from high computational costs, particularly when processing high-resolution images. Recently, state-space models (SSMs) and recurrent neural networks (RNNs) have attracted attention due to their efficiency. However, their performance in image classification tasks remains limited. To address these challenges, this paper introduces VisionGRU, a novel RNN-based architecture designed for efficient image classification. VisionGRU leverages a simplified Gated Recurrent Unit (minGRU) to process large-scale image features with linear complexity. It divides images into smaller patches and progressively reduces the sequence length while increasing the channel depth, thus facilitating multi-scale feature extraction. A hierarchical 2DGRU module with bidirectional scanning captures both local and global contexts, improving long-range dependency modeling, particularly for tasks like semantic segmentation. Experimental results on the ImageNet and ADE20K datasets demonstrate that VisionGRU outperforms ViTs, significantly reducing memory usage and computational costs, especially for high-resolution images. These findings underscore the potential of RNN-based approaches for developing efficient and scalable computer vision solutions. Codes will be available at https://github.com/YangLiu9208/VisionGRU.
\end{abstract} 


\section{Introduction}
\label{sec:intro}


Over the past decade, deep learning has revolutionized the field of computer vision~\cite{lecun2015deep,liu2023cross,wei2023visual}. The advent of Convolutional Neural Networks (CNNs) \cite{krizhevsky2012imagenet} enabled machines to automatically learn complex patterns from pixel data by stacking convolutional layers. This multi-layer feature extraction approach established the foundation for capturing information at various scales and constructing spatial hierarchies. Later advancements, such as Vision Transformers (ViTs) \cite{dosovitskiy2020image, khan2022transformers}, made significant strides by treating images as sequences of patches and utilizing self-attention mechanisms \cite{vaswani2017attention,chou2024metala}. This enabled the model to dynamically focus on different regions of the input data, effectively capturing global dependencies and overcoming the limitations of CNNs in modeling long-range interactions. However, these self-attention mechanisms introduce considerable computational complexity, resulting in substantial resource demands, particularly when processing high-resolution images.

\begin{figure}[t]
  \centering
  \includegraphics[width=1\linewidth]{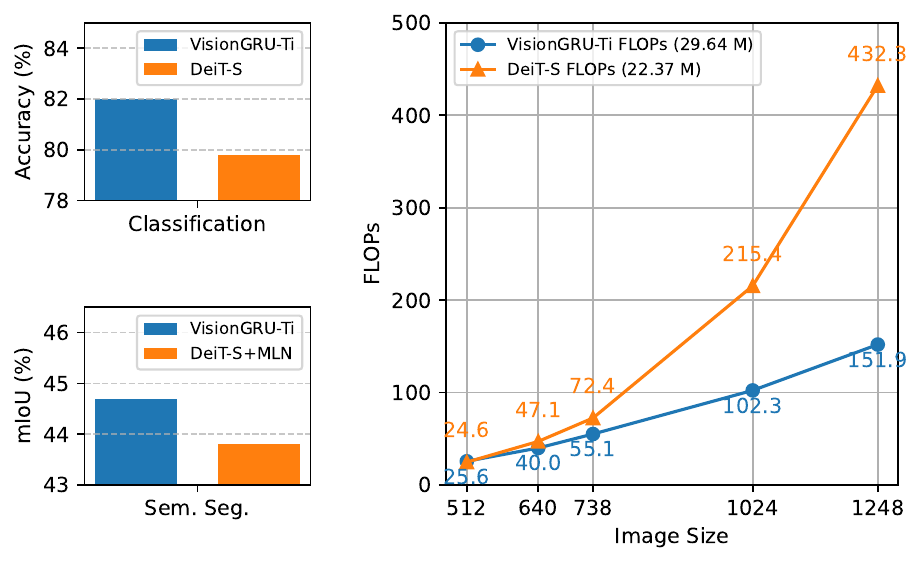} 
  \caption{VisionGRU-Ti achieves higher classification accuracy (82\%) and semantic segmentation mIoU (44.7\%) compared to DeiT-S models. Moreover, VisionGRU-Ti exhibits significantly lower FLOPs at all resolutions, requiring 151.9 GFLOPs at 1248×1248 compared to DeiT-S's 432.3 GFLOPs, highlighting its computational efficiency.}
  \label{fig:gpu_mmc}
\end{figure}

To address this issue, several state-space model (SSM)-based approaches have emerged in recent years, such as Mamba~\cite{gu2022mamba}. These models introduce dynamic state representations, effectively reducing computational complexity and enhancing processing efficiency in visual tasks. However, compared to convolutional and attention-based models, SSMs still exhibit a performance gap, particularly in image classification, where their performance has yet to meet expectations. Similarly, recurrent neural networks (RNNs) have long been essential for modeling sequential data. Nevertheless, they face limitations in computational efficiency and struggle to scale for large-scale image tasks. Recently, researchers have proposed methods for training RNNs using parallel prefix-scan algorithms, with a simplified version of the Gated Recurrent Unit (GRU)\cite{cho2014learning}, known as minGRU\cite{feng2024rnnsneeded}, achieving computational efficiency comparable to that of Mamba.


However, the Recurrent Neural Network (RNN) architecture, originally designed for natural language processing (NLP) tasks, is not inherently suited for image processing. First, the number of pixels in an image is much larger than the number of words in most texts, making it challenging to capture fine-grained features by simply partitioning the image into fixed-size patches. Second, unlike text, images lack an inherent sequential structure, which makes traditional RNNs ill-suited to handle the unordered, two-dimensional nature of image data. Third, training RNNs typically relies on backpropagation through time (BPTT), which involves “unrolling” the network across time steps. For large-scale image data, this unrolling becomes computationally expensive, resulting in slower training speeds and increased memory consumption.

To effectively address the challenges outlined above, this study introduces a novel VisionGRU network architecture. By dividing the image into smaller patches at shallow layers and progressively reducing the sequence length while increasing the number of channels through downsampling operations, VisionGRU outputs multi-scale features while ensuring linear growth in computational complexity. Additionally, a bidirectional scanning mechanism is incorporated into a proposed module, 2DGRU, which modifies the RNN into a time-step reverse-symmetric structure. This enables each local feature to simultaneously integrate global contextual information from both preceding and succeeding segments, thereby enhancing the RNN's capacity to capture long-range dependencies. This approach is particularly well-suited for tasks such as semantic segmentation. Moreover, this modification allows each local feature to incorporate comprehensive context information from both the preceding and succeeding parts, further improving the RNN's ability to capture long-range dependencies, which is advantageous for tasks such as semantic segmentation that rely on intermediate feature map outputs.
As shown in the Fig.~\ref{fig:gpu_mmc}, our model exhibits significant advantages in memory efficiency as the resolution increases.

Our contributions are summarized as follows:
\begin{itemize}
\item To address the high computational and memory overhead of Transformer models in high-resolution image analysis, we introduce the minGRU structure to eliminate the need for BPTT and enables parallel acceleration. Based on this design, we propose VisionGRU to achieve efficient processing of large-scale features with linear computational complexity.

\item To enhance the long-range context modeling capability for 2D features, we design a hierarchical 2DGRU module based on the bidirectional scanning strategy. This approach allows the RNN to effectively capture both local details and global information in 2D spatial domains.

\item Experimental results demonstrate that VisionGRU outperforms mainstream ViT models, such as DeiT, on ImageNet and ADE20K datasets. Specifically, with the same number of parameters, VisionGRU achieves 2.2\% higher accuracy in classification and 0.9\% higher mIoU in segmentation while being 184\% more computationally efficient.

\end{itemize}

\section{Related Work}
\label{sec:related_work}
In recent years, deep learning techniques in image classification have experienced a significant shift, transitioning from Convolutional Neural Networks (CNNs) to Transformer models based on self-attention mechanisms. Each approach offers distinct advantages and limitations regarding feature extraction, computational efficiency, and global information capture. This section provides a comprehensive discussion of these methods and their influence on our work.

\subsection{Convolutional Neural Networks (CNNs) and Their Variants}

Since the introduction of AlexNet \cite{krizhevsky2012imagenet}, which pioneered the use of CNNs in image classification, architectures such as VGG \cite{simonyan2014very}, ResNet \cite{he2016deep}, and DenseNet \cite{huang2017densely} have progressively enhanced feature extraction capabilities. These networks have demonstrated outstanding performance in computational efficiency and local feature capture. However, CNNs face limitations in capturing global dependencies, making it challenging to handle complex image structures and long-range relations. Addressing these limitations often requires deeper networks, which increases computational load and training time.

\subsection{Models Based on Self-Attention Mechanisms}

Initially applied to natural language processing , Transformer\cite{vaswani2017attention} models were later adapted for vision tasks\cite{khan2022transformers}. The Vision Transformer (ViT) \cite{dosovitskiy2020image} utilizes self-attention mechanisms to effectively capture global information from images. However, the computational complexity of ViT scales proportionally with input resolution, leading to significant resource consumption when processing high-resolution images. Variants such as the Swin Transformer \cite{liu2021swin} mitigates this issue through hierarchical self-attention mechanisms and introduce “Patch Merging” to gradually reduce the spatial dimensions of feature maps, enabling multi-scale feature extraction and lower computational complexity. Nevertheless, the high computational resource requirements for complex tasks remain a significant challenge.

\subsection{Vision Models Based on State Space Models (SSM)}

Recently, models based on state-space models (SSMs), such as Mamba \cite{gu2022mamba}, have gained attention for their selective state representation mechanisms. These mechanisms dynamically filter out irrelevant information to focus on the most relevant parts of the input sequence. This approach not only reduces computational overhead but also improves processing efficiency. Inspired by their successful application in natural language processing (NLP) tasks \cite{gu2021efficiently}, many SSM-based vision models have been proposed, such as TranS4mer \cite{nguyen2023trans4mer}, U-Mamba \cite{ma2024umambaenhancinglongrangedependency}, Vision Mamba (Vim) \cite{vim}, and VMamba \cite{li2023vmamba}.

Vision Mamba is a novel vision backbone based on SSMs. It achieves global modeling through bidirectional SSM and has demonstrated efficient visual representation learning capabilities on datasets like ImageNet. Moreover, its bidirectional scanning strategy inspired the design of our 2DGRU module. Despite the attention garnered by SSM-based vision models, their performance in vision tasks like image classification still lags behind other vision Transformer models.
\subsection{Traditional RNNs and Their Limitations}
Traditional Recurrent Neural Networks (RNNs) \cite{elman1990finding} maintain hidden states across time steps, enabling them to capture temporal dependencies, which makes them well-suited for tasks involving time series and natural language processing. However, basic RNNs face challenges, such as vanishing and exploding gradients \cite{bengio1994learning}, when processing long sequences, limiting their ability to model long-term dependencies. To address these issues, enhanced RNN variants like LSTM \cite{hochreiter1997long} and GRU \cite{cho2014learning} incorporate gating mechanisms. Despite these advances, training RNNs using backpropagation through time (BPTT) remains inefficient for long sequences, leading to their gradual replacement by Transformers, which offer superior parallelization capabilities.

\subsection{Efficiently Trainable RNN Variants}

Recently, researchers have proposed novel RNN architectures to improve training efficiency while preserving performance. For example, Orvieto et al. \cite{orvieto2023efficient} and Beck et al. \cite{beck2024complex} introduced efficient RNN variants that use complex diagonal recurrence and exponential gating. The Mamba model, derived from state-space models (SSM), achieves efficient training through input-dependent state transition matrices \cite{gu2022mamba}, demonstrating strong performance in sequence modeling tasks. Additionally, Feng et al. \cite{feng2024minrnn} proposed minLSTM and minGRU, which eliminate the dependence of hidden states on input, forget, and update gates, enabling training with parallel prefix scan algorithms. This simplification reduces the parameter count and significantly improves training speed, with minGRU being approximately 175 times faster than traditional GRU on sequences of length 512. minGRU achieves this efficiency by: (1) removing the dependency on previous hidden states for parallel computation, and (2) eliminating candidate state range constraints (tanh) to ensure temporal independence in outputs. These changes allow minGRU to maintain performance comparable to Mamba and other modern sequence models while achieving linear complexity, which is crucial for long-sequence tasks due to significant resource savings. Like minGRU, some linear attention-based variants also aim to improve training parallelization. Although these approaches have made notable progress in training efficiency and performance, minGRU’s design strikes an effective balance between reduced parameters and preserved performance.

\section{Methodology}
\label{method}
This section presents the design and implementation details of the VisionGRU model. The core innovation of VisionGRU lies in combining the computational advantages of Recurrent Neural Networks (RNNs) with the ability to process 2D features, enabling efficient global feature capture through the 2DGRU module. The model’s overall architecture and key modules are outlined step by step below.

\begin{figure*}[t]
  \centering
  \includegraphics[width=0.9\linewidth]{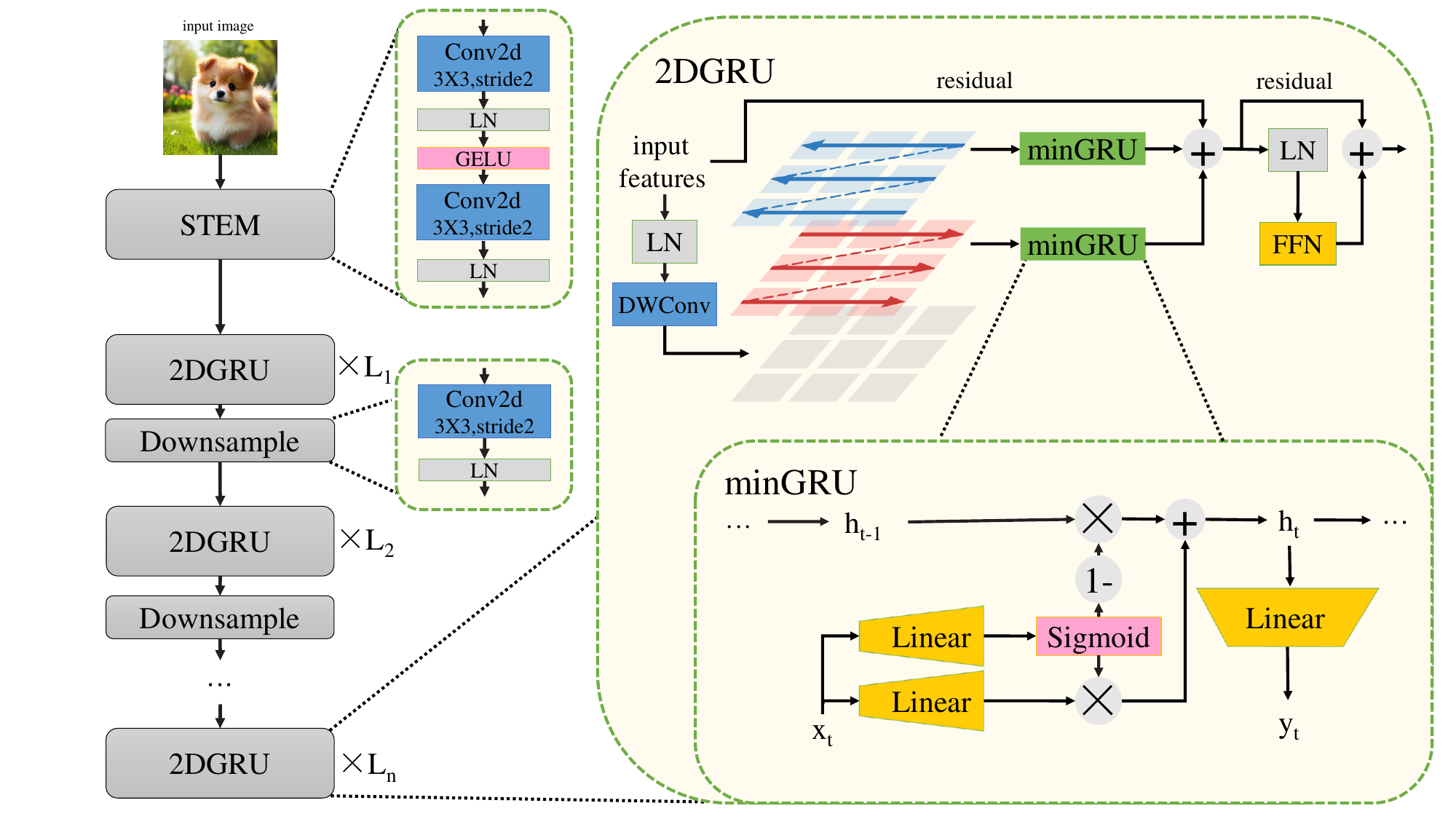} 
  \caption{Overview of the VisionGRU model. It integrates the computational strengths of RNNs and CNNs using a hierarchical 2DGRU module with a bidirectional scanning strategy for efficient global feature capture.}
  \label{fig:model-architecture}
\end{figure*}

\subsection{Preliminary}
\subsubsection{RNN Theory}
Recurrent Neural Networks (RNNs) are network architectures designed to capture temporal information and continuity by propagating states across time steps, thereby maintaining connections between different points in time. The basic RNN formula is as follows:
\begin{equation}
\mathbf{h}_t = \phi(\mathbf{W}_h \mathbf{h}_{t-1} + \mathbf{W}_x \mathbf{x}_t)
\end{equation}

where \(\mathbf{h}_t\) represents the hidden state at the current time step, \(\mathbf{W}_h\) and \(\mathbf{W}_x\) are learnable parameter matrices, and \(\phi\) is an activation function. Basic RNNs face gradient vanishing and exploding problems when handling long sequences, limiting their ability to model long-term dependencies.
\subsubsection{GRU Unit}
To address the gradient vanishing problem of RNNs, Gated Recurrent Unit (GRU) was introduced. GRU utilizes update and reset gates to regulate information flow, improving the ability to capture long-term dependencies. The GRU equations are as follows:
\begin{equation}
z_t = \sigma\left(\mathbf{W}_z \mathbf{x}_t + \mathbf{U}_z \mathbf{h}_{t-1}\right)
\end{equation}
\begin{equation}
r_t = \sigma\left(\mathbf{W}_r \mathbf{x}_t + \mathbf{U}_r \mathbf{h}_{t-1}\right)
\end{equation}
\begin{equation}
\tilde{\mathbf{h}}_t = \tanh\left(\mathbf{W}_h \mathbf{x}_t + \mathbf{U}_h \left(r_t \odot \mathbf{h}_{t-1}\right)\right)
\end{equation}
\begin{equation}
\mathbf{h}_t = \left(1 - z_t\right) \odot \mathbf{h}_{t-1} + z_t \odot \tilde{\mathbf{h}}_t
\end{equation}
where \(\sigma\) denotes the Sigmoid function, \(\odot\) represents element-wise multiplication, \(z_t\) is the update gate that controls the carryover of the previous hidden state \(\mathbf{h}_{t-1}\), \(r_t\) is the reset gate that modulates the influence of \(\mathbf{h}_{t-1}\) in computing the candidate hidden state \(\tilde{\mathbf{h}}_t\), and \(\mathbf{h}_t\) is the final hidden state at time \(t\), balancing information from \(\mathbf{h}_{t-1}\) and \(\tilde{\mathbf{h}}_t\) based on \(z_t\). 

GRUs effectively mitigate the vanishing gradient problem, but their reliance on backpropagation through time (BPTT) can lead to inefficiencies in handling very long sequences, limiting parallel computation.

\subsubsection{minGRU Unit}
The minGRU is a simplified version of the traditional GRU \cite{cho2014learning} that achieves efficient parallel training by removing dependencies between hidden states. Specifically, the minGRU eliminates the reset gate from the traditional GRU and removes the dependence of the update gate and candidate hidden state on previous hidden states, while also excluding the tanh activation function to ensure time-scale independence of the output. The minGRU update equation is:
\begin{align}
    h_t &= (1 - z_t) \odot h_{t-1} + z_t \odot \tilde{h}_t \\
    z_t &= \sigma(\text{Linear}_{d_h}(x_t)) \\
    \tilde{h}_t &= \text{Linear}_{d_h}(x_t)
\end{align}
where \( z_t \) represents the update gate, and \( \text{Linear}(x_t) \) is a linear transformation of the input. These modifications reduce the number of parameters and enhance training efficiency \cite{feng2024minrnn}.

\subsection{VisionGRU Architecture}
The overall architecture of VisionGRU is shown in Fig~\ref{fig:model-architecture}. The input image first undergoes initial feature extraction through a shallow convolutional network, with positional embeddings added to preserve spatial information. These features are then processed through a series of 2DGRU modules. Inspired by the Swin model, we incorporate downsampling layers to progressively reduce the spatial dimensions of the feature maps, thereby aggregating high-level semantic information and reducing computational costs. Specifically, the feature map processed by the stem and enhanced with positional encoding serves as the input to the first 2DGRU, denoted as \( F \in \mathbb{R}^{H/4 \times W/4 \times D} \), where \( D \) is the hidden dimension. Subsequently, these feature maps are treated as a series of tokens in the 2D space, which are fed in both forward and reverse raster scanning orders into the 2DGRU blocks in parallel. Downsampling is applied at specific points to further reduce the spatial dimensions and expand the channel width, as detailed in the experimental section. The final output feature \( F_{\text{out}} \in \mathbb{R}^{H/32 \times W/32 \times D\times4} \) is utilized for classification.


\subsection{2DGRU Module}

The 2DGRU module is the core component of the VisionGRU model, designed to model global spatial dependencies through a bidirectional scanning strategy. The input feature map \( F_{\text{in}} \) is first normalized using layer normalization, then processed through depthwise convolution and computed using minGRU units. The output features from different directions are aggregated at corresponding positions and passed through a residual connection, followed by an FFN module to produce the final output feature map.

\begin{algorithm}[h]
\caption{2DGRU Block Process}
\textbf{Input:} Input feature map $F_{\text{in}} : (B, H, W, C)$\\
\textbf{Output:} Output feature map $F_{\text{out}} : (B, H, W, C)$

1:  /* Normalize the input feature map */\\
2:  $F_{\text{norm}} : (B, H, W, C) \leftarrow \text{Norm}(F_{\text{in}})$\\
3:  /* Apply depthwise convolution */\\
4:  $F_{\text{conv}} : (B, H, W, C) \leftarrow \text{DepthwiseConv}(F_{\text{norm}})$\\
5:  /* Process bidirectionally */\\
6:  For each path $p \in \{\text{Paths}\}$:\\
7:  \quad $x_p : (B, H \cdot W , C) \leftarrow \text{minGRU}_p(F_{\text{conv}} \text{ along } p)$\\
8:  End for\\
9:  /* Aggregate results */\\
10: $x \leftarrow \sum_{p} x_p$\\
11: /* Add residual connection */\\
12: $F_{\text{out}} : (B, H, W, C) \leftarrow x + F_{\text{in}}$\\
\Return $F_{\text{out}}$
\vspace{0.5em} 
\end{algorithm}

To further illustrate the parallelism within the 2DGRU module, we expand the recurrence formula for each direction of the minGRU. Assuming the initial hidden state for each minGRU is a zero vector, this assumption eliminates dependencies on previous hidden states, simplifying computation. The output feature at position \((i, j)\), denoted as \( f_{(i, j)}^{\text{out}} \), is expressed as:

\begin{equation}
\begin{aligned}&f_{\left( i,j\right) }^{out}=f^{in}_{\left (i, j\right )} + \sum ^{M-1}_{m=0}\left(y_{\rightarrow,m} + y_{\leftarrow,m}\right)
\end{aligned}
\end{equation}
\begin{equation}
y_{\rightarrow,m} =
\begin{cases}
\prod\limits_{k=m+1}^{L_{i,j}} \left( 1-z_{k,\rightarrow} \right) \odot z_{m,\rightarrow} \odot h_{m,\rightarrow}& m\leq L_{i,j} \\
0 & m > L_{i,j}
\end{cases} 
\end{equation}
\begin{equation}
\begin{aligned}
y_{\leftarrow,m} =
\begin{cases}
\prod\limits_{k=L_{i,j}}^{m-1} \left( 1-z_{k,\leftarrow} \right) \odot z_{m,\leftarrow} \odot h_{m,\leftarrow}& m \geq L_{i,j} \\
0 & m < L_{i,j}
\end{cases}
\end{aligned}
\end{equation}

Where:

\begin{itemize}
    \item $f_{(i,j)}^{in}$: The input feature at position $(i, j)$.
    \item $f_{(i,j)}^{out}$: The output feature at position $(i, j)$.
    \item $M$: The total length of the scan sequence.
    \item $L_{i,j}$: The index position of the feature at $(i,j)$ in the scan sequence.
    \item $z_{k,\rightarrow}, z_{m,\rightarrow}$: Forward update gates, which determine how new candidate features are accepted in the forward direction.
    \item $h_{m,\rightarrow}$: The forward candidate hidden state.
    \item $z_{k,\leftarrow}, z_{m,\leftarrow}$: Backward update gates.
    \item $h_{m,\leftarrow}$: The backward candidate hidden state.
    \item $\displaystyle \prod_{k=m+1}^{L_{i,j}} \bigl(1 - z_{k,\rightarrow}\bigr)$: The residual factor in the forward path, accumulating non-updated (i.e., $1 - z$) terms from step $m+1$ to $L_{i,j}$.
    \item $\displaystyle \prod_{k=L_{i,j}}^{m-1} \bigl(1 - z_{k,\leftarrow}\bigr)$: The residual factor in the backward path, accumulating non-updated terms from step $L_{i,j}$ to $m-1$.
\end{itemize}

Thus, each output depends solely on the input features, enabling efficient parallelization.

\subsection{Details}  
The VisionGRU is implemented in two configurations: Base and Tiny. The Base configuration consists of 21 2DGRU blocks, with three downsampling layers inserted. These downsampling layers divide the network into four distinct stages, containing [2, 2, 15, 2] 2DGRU blocks respectively. For the Tiny configuration, the network is adjusted to be lightweight, consisting of a total of 14 2DGRU blocks. Similar to the Base configuration, three downsampling layers are inserted, dividing the network into four stages with the block distribution set to [2, 2, 8, 2].  Both configurations utilize the same principles for feature extraction and downsampling, but the Tiny configuration is optimized for scenarios requiring lower computational costs, making it suitable for lightweight applications. 
\section{Experiment}  
\label{experiment}  

\subsection{Image Classification}  
We conducted benchmark tests on the ImageNet-1K dataset \cite{deng2009imagenet}, which contains 1.28 million training images and 50,000 validation images across 1,000 classes. Our training setup primarily followed DeiT \cite{touvron2021training}. The input images were resized to a resolution of $224 \times 224$, and the final $7 \times 7$ feature maps were average-pooled and passed through a fully connected layer for classification. We used the AdamW optimizer \cite{loshchilov2017decoupled} with a momentum of 0.9 and a weight decay of 0.05. Training was performed on eight A800 GPUs for a total of 300 epochs. We used cosine scheduling with a learning rate scaling rule \(1 \times 10^{-3} \times \text{batch size} / 1,024\). Additionally, we applied Exponential Moving Average (EMA) \cite{polyak1992acceleration} with a decay rate of 0.9999. 

\begin{table}[t]  
\centering  \renewcommand\arraystretch{1.5}
\caption{Comparison of model architectures in terms of parameter count, input resolution, and ImageNet top-1 accuracy.} 
\begin{tabular}{|c|c|c|c|}  
\hline  
\textbf{Method} & \textbf{\#Params} & \textbf{Image Size} & \textbf{Top-1 (\%)} \\  
\hline  
\multicolumn{4}{|l|}{\textbf{ConvNets}} \\  
\hline  
VGG11 & 132.8M & $224\times 224$ & 69.1 \\  
ResNet-18 \cite{he2016deep} & 12M & $224\times 224$ & 69.8 \\  
ResNet-50 \cite{he2016deep} & 25M & $224\times 224$ & 76.2 \\  
ResNet-101 \cite{he2016deep} & 45M & $224\times 224$ & 77.4 \\  
ResNet-152 \cite{he2016deep} & 60M & $224\times 224$ & 78.3 \\  
ResNeXt50-32 $\times$ 4d \cite{xie2017aggregated} & 25M & $224\times 224$ & 77.6 \\  
RegNetY-4GF \cite{radosavovic2020designing} & 21M & $224\times 224$ & 80.0 \\  
\hline  
\multicolumn{4}{|l|}{\textbf{Transformers}} \\  
\hline  
ViT-B/16 \cite{dosovitskiy2020image} & 86M & $384\times 384$ & 77.9 \\  
ViT-L/16 \cite{dosovitskiy2020image} & 307M & $384\times 384$ & 76.5 \\  
DeiT-Ti \cite{touvron2021training} & 6M & $224\times 224$ & 72.2 \\  
DeiT-S \cite{touvron2021training} & 22M & $224\times 224$ & 79.8 \\  
DeiT-B \cite{touvron2021training} & 86M & $224\times 224$ & 81.8 \\  
\hline  
\multicolumn{4}{|l|}{\textbf{State-Space Models (SSMs)}} \\  
\hline  
S4ND-ViT-B \cite{gu2022mamba} & 89M & $224\times 224$ & 80.4 \\  
Vim-Ti \cite{li2023vmamba} & 7M & $224\times 224$ & 76.1 \\  
Vim-S \cite{li2023vmamba} & 26M & $224\times 224$ & 80.5 \\  
\hline  
\multicolumn{4}{|l|}{\textbf{RNN Models}} \\  
\hline  
 \textbf{VisionGRU-Ti} & \textbf{30M} & \textbf{$224\times 224$} & \textbf{82.0} \\  
 \textbf{VisionGRU-B} & \textbf{86M} & \textbf{$224\times 224$} & \textbf{83.1} \\  
\hline  
\end{tabular}  
\label{tab:model_comparison}  
\end{table}  

Table~\ref{tab:model_comparison} highlights the significant performance gains achieved by VisionGRU when compared to diverse model families, including convolution-based (e.g., ResNet), attention-based (e.g., DeiT), and state-space models (e.g., Vim). Specifically, VisionGRU-Ti attains an 82.0\% top-1 accuracy, surpassing both the similarly lightweight DeiT-Ti (72.2\%) and Vim-Ti (76.1\%). Moreover, with a comparable parameter budget to DeiT-B, VisionGRU-B achieves 83.1\% top-1 accuracy—an improvement of 1.3 percentage points over DeiT-B’s 81.8\%. Notably, these performance gains emerge despite VisionGRU using a straightforward hierarchical RNN structure with minimal architectural optimizations.

A primary reason for these improvements lies in VisionGRU’s bidirectional 2DGRU module, which ensures that each spatial location aggregates information from preceding and succeeding regions in both directions. This global feature fusion alleviates common issues in standard RNNs, where long-range dependencies often vanish through sequential steps. Additionally, VisionGRU’s hierarchical downsampling design allows the network to capture multiple feature scales: finer spatial details in earlier layers and more abstract, high-level representations in deeper layers. Consequently, the model balances local detail preservation and global context integration, yielding robust classification performance. Our experimental findings also suggest that the GRU’s gating mechanism helps filter out redundant information and focus on salient features—an advantage over pure attention-based methods, which can become computationally expensive when modeling high-resolution inputs.

\begin{table}[t]  
\centering  \renewcommand\arraystretch{1.5}
\setlength{\tabcolsep}{4pt}
\caption{Semantic segmentation results on the ADE20K validation set.}  
\begin{tabular}{|c|c|c|c|c|}  
\hline  
\textbf{Method} & \textbf{Backbone} & \textbf{Image Size} & \textbf{\#Params} & \textbf{mIoU (\%)} \\  
\hline  
DeepLab v3+ & ResNet-101 & $512\times 512$ & 63M & 44.1 \\  
UperNet & ResNet-50 & $512\times 512$ & 67M & 41.2 \\  
UperNet & Vim-Ti & $512\times 512$ & 13M & 41.0 \\  
UperNet & Vim-S & $512\times 512$ & 46M & 44.9 \\  
UperNet & Swin-Ti & $512\times 512$ & 60M & 44.5 \\  
UperNet & DeiT-S + MLN & $512\times 512$ & 58M & 43.8 \\  
\textbf{UperNet} & \textbf{VisionGRU-Ti} & \textbf{$512\times 512$} & \textbf{60M} & \textbf{44.7} \\  
\hline  
\end{tabular}  
\label{tab:semantic_segmentation_results}  
\end{table}  

\begin{figure*}[t]
  \centering
  \includegraphics[width=1\linewidth]{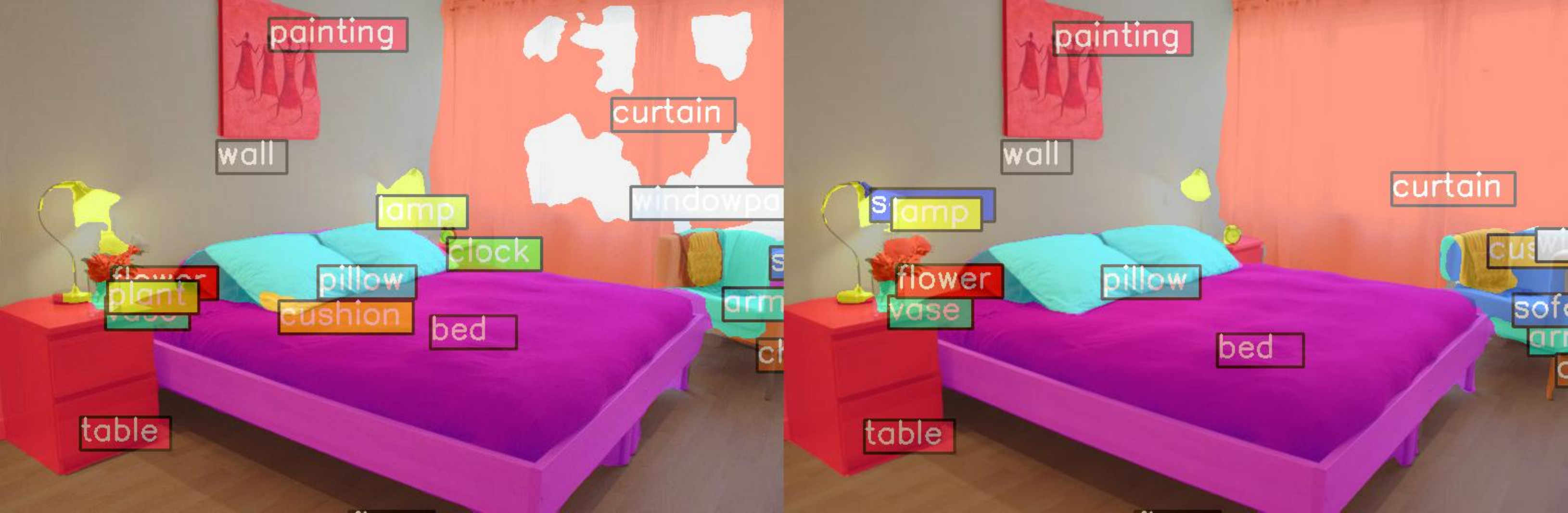} 
  \caption{The semantic segmentation example from the ADE20K validation set. The left image shows the segmentation result of Swin, while the right image shows the result of our VisionGRU. }
  \label{fig:ade_cmp}
\end{figure*}

\subsection{Semantic Segmentation}  

To further test VisionGRU’s generalization ability, we integrated it into a UperNet\cite{xia2020upernet} framework for semantic segmentation experiments on ADE20K. As shown in Table~\ref{tab:semantic_segmentation_results}, UperNet equipped with VisionGRU-Ti achieves a mean Intersection-over-Union (mIoU) of 44.7\%, outperforming DeiT-S + MLN (43.8\%) and tying closely with more specialized architectures such as Swin-Ti (44.5\%). This 0.9\% improvement over the DeiT backbone underscores VisionGRU’s capability to extract stronger contextual cues, which is crucial for pixel-level tasks. As shown in Figure~\ref{fig:ade_cmp}, our model, leveraging the 2DGRU's ability to capture long-range dependencies, is more effective at handling segmentation tasks, such as identifying the curtain in the image, which lacks distinct local features and requires global contextual information for accurate predictions. In contrast, Swin, due to the quadratic complexity of self-attention, divides the image into windows, resulting in limited effective information within each window, which hampers its ability to make accurate predictions in such cases.

Unlike conventional attention-based models, VisionGRU’s recurrent design leverages bidirectional 2D scans to encode both local and global spatial information in a memory-efficient way. This advantage becomes increasingly relevant at higher input resolutions, where self-attention can be prohibitively costly. Furthermore, our experiments indicate that VisionGRU’s parallel-friendly architecture—enabled by the minGRU-like structure and prefix-scan training algorithms—makes it straightforward to scale to large images without excessive computation. Consequently, VisionGRU not only shows strong classification accuracy but also delivers competitive results on dense prediction tasks, suggesting its versatility across diverse vision applications.

\subsection{Discussion and Practical Implications} Overall, VisionGRU's improvements in both image classification and semantic segmentation highlight the effectiveness of combining hierarchical RNN structures with a bidirectional scanning strategy. By integrating multi-scale feature extraction, efficient gating mechanisms, and global spatial context modeling, VisionGRU demonstrates that recurrent networks can remain competitive—and even surpass state-of-the-art transformer-based methods—when specifically redesigned for 2D data processing. From a practical perspective, VisionGRU's linear computational complexity and reduced FLOPs at higher resolutions (Figure~\ref{fig:gpu_mmc}) present a promising alternative for tasks that require real-time processing or large-scale deployment. We believe these findings not only confirm the resurgence of RNN-based designs in computer vision but also open new directions for more works \cite{liu2021semantics,liu2022tcgl,ma2023enhanced,liu2022causal,yan2023skeletonmae,zhu2022hybrid,chen2023cross,liu2024aligning,song2024towards,liu2024fine,zheng2024diversity} (e.g., video analysis, embodied AI, spatial-temporal prediction, medical image analysis and recommendation system) in efficient model architectures.

\section{Conclusion}  
\label{conclusion}  

This paper introduces VisionGRU, a novel vision backbone architecture that capitalizes on the strengths of RNNs through the incorporation of 2DGRU modules, enabling efficient feature learning and global dependency modeling. The hierarchical structure of VisionGRU, coupled with parallel bidirectional minGRU units, facilitates effective multi-scale feature extraction while maintaining low computational complexity. Extensive experiments on the ImageNet-1K dataset demonstrate that VisionGRU outperforms traditional convolutional neural network architectures in both accuracy and efficiency. The model not only reduces computational overhead significantly but also sets new benchmarks for visual tasks, achieving state-of-the-art performance. Future research can leverage these findings to further optimize the architecture and expand the applicability of VisionGRU to other complex vision tasks.

\bibliographystyle{IEEEbib}
\bibliography{main}

\end{document}